\documentclass[12pt]{article}

\usepackage{sbc-template}
\usepackage{graphicx,url}
\usepackage{subcaption}
\usepackage[utf8]{inputenc}
\usepackage[english]{babel}
\usepackage{amsmath}
\usepackage{multirow}
\usepackage{booktabs} 
\usepackage{arydshln}
\usepackage[table]{xcolor}

\usepackage{colortbl}
\title{Risk-Aware Robust Learning: Reducing Clinical Risk under Label Noise in Medical Image Classification}

\author{Maycon R. S. Pereira, Filipe R. Cordeiro}

\address{
  Visual Computing Lab, Departamento de Computação, \\ Universidade Federal Rural de Pernambuco (UFRPE), Brasil\\
  \{sromario577@gmail.com, filipe.rolim@ufrpe.br\}  
}

\begin{document} 

\maketitle

\begin{abstract}
Noisy labels are a pervasive challenge in medical image classification, where annotation errors arise from inter-observer variability and diagnostic ambiguity. Although several noise-robust learning methods have been proposed, their evaluation predominantly relies on accuracy-oriented metrics, overlooking the clinical implications of asymmetric error costs. In medical diagnosis, a false negative (missed disease) carries substantially higher consequences than a false positive (false alarm), as delayed treatment can directly impact patient outcomes. In this work, we investigate whether noise-robust training methods preserve clinical safety under label noise. We conduct a systematic risk-aware evaluation of the state-of-the-art noise-robust methods Co-teaching, DivideMix, UNICON, and a GMM-based filtering approach on binarized DermaMNIST and PathMNIST datasets under clean and label noise rates of  20\%, and 40\%. Beyond balanced accuracy, we adopt a cost-sensitive Global Risk formulation that explicitly penalizes false negatives. Our analysis reveals that the robustness of state-of-the-art methods does not guarantee clinical safety. 
Furthermore, we demonstrate that integrating cost-sensitive optimization into noise-robust training significantly reduces clinical risk, while mantaining model utility. These findings demonstrate that noise-robust learning must be evaluated through a clinical risk lens, and that combining robust training with cost-sensitive optimization can meaningfully reduce risk in noisy-label medical imaging scenarios.

\end{abstract}
     


\section{Introduction}

The application of Deep Learning to medical image classification has achieved remarkable diagnostic performance, reaching accuracy levels comparable to human specialists in tasks such as skin lesion classification and histopathological analysis~\cite{chan2020deep}. However, these models rely on large-scale annotated datasets that are frequently subject to label noise arising from inter-observer variability, annotator fatigue, and inherent ambiguity in medical cases~\cite{Frenay2014, Cordeiro2023}.

To address this challenge, several noise-robust learning methods have been proposed, employing strategies such as sample selection based on loss values~\cite{Han2018, Jiang2018}, Gaussian Mixture Models (GMM) to separate clean from noisy samples~\cite{Li2020, Karim2022}, and semi-supervised learning techniques~\cite{Li2020}. While these approaches demonstrate improvements under standard evaluation protocols, they are typically assessed using symmetric metrics such as accuracy or balanced accuracy, which treat all misclassification errors equally.

This evaluation paradigm is inadequate for medical diagnosis, where misclassification costs are inherently asymmetric~\cite{Elkan2001, Ling2006}. A false negative, failing to detect a malignant condition, may delay treatment with potentially life-threatening consequences, whereas a false positive typically leads to additional confirmatory tests~\cite{Scholz2024}. This cost asymmetry fundamentally alters how model performance should be assessed: a classifier that achieves high balanced accuracy but misses malignant cases poses greater clinical danger than one with lower accuracy but fewer missed diagnoses. As demonstrated by Haimerl and Reich~\cite{haimerl2025}, 
neglecting cost asymmetries when evaluating ML-based classifiers  for medical devices can increase the resulting expected risk by  up to 196\% when the ratio between false negative and false  positive costs changes by a factor of~10. This finding  underscores the inadequacy of symmetric evaluation metrics for  clinical deployment scenarios.

Despite this well-known asymmetry, existing studies on learning with noisy labels have not systematically investigated how noise-robust methods affect clinical risk. This raises a critical question: \textbf{do noise-robust training methods preserve clinical safety under label noise, or do they inadvertently amplify diagnostic risk by shifting the false negative/false positive balance?}


    
    


To investigate this, we conduct a systematic risk-aware evaluation of state-of-the-art noise-robust methods under controlled noise levels. Furthermore, we examine whether integrating cost-sensitive optimization during training can meaningfully reduce clinical risk. Our contributions are the following:

\begin{itemize}
\item We provide a systematic clinical risk analysis of noise-robust learning methods in medical image classification, with different levels of label noise rate.
\item We show the importance of assessing robust training methods considering the risk analysis, instead of only acuracy-based metrics for medical images.
\item We demonstrate that integrating cost-sensitive optimization into noise-robust training significantly reduces clinical risk under label noise.
\end{itemize}

We evaluate our approach on two binary classification tasks derived from MedMNIST~\cite{medmnist}: DermaMNIST (skin lesion classification) and PathMNIST (colorectal tissue classification), under 0\%, 20\% and 40\% noise rates. 
We focus on binary classification (malignant vs.\ benign) for two complementary reasons. First, binary screening represents the most critical decision point in clinical diagnostic pipelines. 
Second, the binary formulation enables a clean and clinically interpretable risk analysis with only two cost parameters (false negative and false positive), avoiding the combinatorial complexity of specifying a full cost matrix for multi-class problems, where pairwise misclassification costs are often difficult to justify clinically.

Our findings demonstrate that robustness and safety are distinct properties and that risk-aware optimization is essential for safe deployment of medical systems trained under noisy annotations.

\section{Background}


Deep neural networks can easily memorize incorrect labels during training, leading to poor generalization~\cite{Arpit2017, Zhang2021}. To mitigate this, several families of approaches have been proposed in the literature~\cite{Song2022, Carneiro2024}.

Sample selection methods exploit the observation that clean samples are learned earlier than noisy ones. Co-teaching~\cite{Han2018} trains two networks simultaneously, each selecting small-loss samples to update its peer. GMM-based approaches model per-sample loss distributions to separate clean from noisy samples~\cite{Li2020}. DivideMix~\cite{Li2020} combines GMM-based filtering with semi-supervised learning. UNICON~\cite{Karim2022} extends this framework by incorporating uniform sample selection and contrastive learning. While these methods improve robustness under symmetric evaluation metrics, their implications for asymmetric clinical risk remain underexplored.

In medical diagnosis, misclassification costs are inherently asymmetric: false negatives (missing a disease) typically lead to delayed treatment and worse patient outcomes, while false positives result in additional tests or temporary anxiety~\cite{Elkan2001, Ling2006}. Cost-sensitive learning addresses this asymmetry through loss reweighting based on class-specific costs~\cite{Khan2018}, decision threshold adjustment~\cite{Collell2018}, and cost-sensitive evaluation metrics~\cite{Scholz2024}.

To the best of our knowledge, no prior work has systematically analyzed how noise-robust learning methods affect clinical risk in medical image classification, nor assessed how cost-sensitive loss interacts with these methods under label noise.

\section{Methodology}

\subsection{Problem Formulation}

Let $\mathcal{D} = \{(x_i, y_i)\}_{i=1}^{N}$ be a training dataset where $x_i$ is a medical image and $y_i \in \{0, 1\}$ is its binary label, with $y_i = 1$ denoting the malignant (positive) class and $y_i = 0$ the benign (negative) class. In the presence of label noise, the observed label $y_i$ may differ from the hidden true label $\hat{y}_i$ due to annotation errors. We consider symmetric label noise~\cite{Frenay2014}, where each training label is independently flipped with probability $\eta$, i.e., $P(\tilde{y} \neq y) = \eta$, where $\tilde{y}$ denotes the corrupted label. The noise is symmetric (uniform) in that corruption is equally likely across classes. The noise is considered asymmetric if the noise occurs between specific classes.  For the binary classification problem, symmetric and asymmetric noises are the same.

\subsection{Datasets}

We use DermaMNIST and PathMnist, two datasets from the MedMNIST collection~\cite{Yang2023}, converted to binary classification tasks. All images are resized to 64$\times$64 pixels.

\textbf{DermaMNIST} consists of 10,015 dermatoscopic images of pigmented skin lesions from the HAM10000 dataset. The original 7 diagnostic categories are binarized by grouping melanoma, basal cell carcinoma, and actinic keratosis as \textit{malignant}, while melanocytic nevus, benign keratosis, vascular lesion, and dermatofibroma are grouped as \textit{benign}.

\textbf{PathMNIST} contains 107,180 colon pathology image patches from the NCT-CRC-HE-100K dataset. The original 9 tissue types are binarized by grouping colorectal adenocarcinoma epithelium and cancer-associated stroma as \textit{malignant}, with the remaining seven classes grouped as \textit{benign}.


Table~\ref{tab:dataset_stats} summarizes the class distribution after binarization. Both datasets exhibit class imbalance representative of real-world screening populations, with malignant cases comprising approximately 20--25\% of samples.

\begin{table}[h]
\centering
\caption{Dataset statistics of DermaMNIST and PathMNIST after binarization. Original multi-class labels are grouped into clinically meaningful binary categories.}
\label{tab:dataset_stats}
\begin{tabular}{l|cc|cc}
\hline
\multirow{2}{*}{Split} & \multicolumn{2}{c|}{DermaMNIST} & \multicolumn{2}{c}{PathMNIST} \\
 & Benign & Malignant & Benign & Malignant \\
\hline
Train & 5,641 (80.5\%) & 1,366 (19.5\%) & 67,710 (75.2\%) & 22,286 (24.8\%) \\
Val & 807 (80.5\%) & 196 (19.5\%) & 7,527 (75.2\%) & 2,477 (24.8\%) \\
Test & 1,613 (80.4\%) & 392 (19.6\%) & 5,526 (77.0\%) & 1,654 (23.0\%) \\
\hline
\end{tabular}
\end{table}

We apply symmetric label noise to the training set at rates $\eta \in \{0\%, 20\%, 40\%\}$. For each noise rate, we independently flip the label of each training sample with probability $\eta$. The validation and test sets remain clean to ensure reliable evaluation.



\subsection{Evaluated Methods}

We evaluate the noise-robust learning approaches Gaussian Mixture Model (GMM) Filter~\cite{Li2020}, Co-teaching~\cite{Han2018}, DivideMix~\cite{Li2020} and UNICON~\cite{Karim2022},  besides baseline. Baseline trains a standard classifier with cross-entropy loss, without any noise-handling mechanism.

Gaussian Mixture Model (GMM) Filterfits a two-component Gaussian Mixture Model to the per-sample loss distribution at each epoch, selecting samples with high probability of belonging to the low-loss (clean) component using a threshold of 0.5. It is present in robust learning solutions, as in ~\cite{Li2020, anne}, but here it is evaluation as a filtering adition to the baseline, to understand its impact on clinical risk.  

Co-teaching~\cite{Han2018} trains two networks simultaneously where each selects small-loss samples to train its peer, exploiting different learning dynamics across networks. It is also a standard solution for noisy label filtering, which is based on the agreement of two models. 

DivideMix~\cite{Li2020} combines GMM-based sample division and co-teaching with semi-supervised learning, using clean samples for supervised training and treating noisy samples as unlabeled data through MixMatch~\cite{Berthelot2019}. UNICON~\cite{Karim2022} extends DivideMix with uniform sample selection across classes and contrastive learning for more robust feature representations under label noise.


For each method, we also evaluate a variant with Cost-Sensitive (CS) Loss~\cite{araf2024cost}, which reweights the cross-entropy loss based on class-specific costs:
\begin{equation}
\mathcal{L}_{CS} = -\sum_{i=1}^{N} w_{y_i} \log(\hat{p}_{y_i})
\end{equation}

\noindent where $w_{y_i}$ is the cost weight for class $y_i$. The variants of Baseline, GMM Filter, Co-teaching, DivideMix and UNICON, with the cost-sensitive loss, are named Baseline+CS, GMM Filter+CS, DivideMix+CS and UNICON+CS, respectively. 


\subsection{Evaluation Metrics}

We evaluate the methods using both standard binary classification  and clinical risk metrics. For the classification metrics we used Balanced Accuracy, (BAC), Sensitivity and Specificity. 

Sensitivity measures the proportion of malignant cases correctly identified by the equation $TP / (TP + FN)$, with $TP$ and $FN$ representing the number of true positives and false negatives, respectively.

Specificity measures the proportion of benign cases correctly identified by $TN / (TN + FP)$, where $TN$ represent the true negatives.

BAC is the arithmetic mean of sensitivity and specificity, providing fair evaluation under class imbalance:
\begin{equation}
\text{BAC} = \frac{1}{2}\left(\frac{TP}{TP + FN} + \frac{TN}{TN + FP}\right)
\end{equation}

Following the Global Risk formulation proposed by 
Haimerl and Reich~\cite{haimerl2025}, which demonstrates that 
risk-based considerations must be integrated into the evaluation 
of Machine Learning (ML)-based classifiers for medical applications, we adopt a 
cost-sensitive risk metric defined as:
\begin{equation}
  \text{Risk} = \frac{C_{\mathit{FN}} \cdot \mathit{FN} + C_{\mathit{FP}} \cdot \mathit{FP}}{N}
  \label{eq:risk}
\end{equation}
where  $N$ is the total number of test samples,  and $C_{FN}$, $C_{FP}$ are the misclassification costs assigned  to each error type. This formulation captures the expected 
misclassification cost per sample and is consistent with the 
expected risk framework in decision 
theory~\cite{Elkan2001, Ling2006}.

We define the  two clinical risk evaluation scenarios:
\begin{itemize}
  \item \textbf{Risk~I} ($C_{FN} = 1$, $C_{FP} = 1$): Equal error costs, corresponding to the 
    standard assumption in most machine learning evaluation, 
    where all misclassifications are treated symmetrically.
  \item \textbf{Risk~II} ($C_{FN} = 20$, $C_{FP} = 1$, ): Asymmetric costs reflecting clinical 
    priorities where missing a malignant case is substantially 
    more costly than a false alarm.
\end{itemize}

The adoption of $C_{FN} = 20$ for Risk~II is a representative value for serious disease screening scenarios, as there is no universally agreed-upon cost ratio in the 
literature, as it depends on context. We additionally report Risk~I ($\lambda = 1$) to enable direct 
comparison with the equal-cost assumption, allowing practitioners  to assess how the choice of cost ratio affects method ranking. Lower risk values indicate safer models. Risk~II specifically  penalizes methods that sacrifice sensitivity for specificity



\subsection{Experimental Setup}

We use ResNet-18~\cite{He2016} pretrained on ImageNet as the backbone for all experiments. Models are trained for 200 epochs using SGD with momentum 0.9, learning rate 0.01 with cosine annealing, and batch size 64. Data augmentation includes random cropping  and random horizontal flipping. Methods requiring warmup (GMM Filter, DivideMix, UNICON) use 10 warmup epochs with standard cross-entropy loss. 

For cost-sensitive loss, as defined in Equation~\ref{eq:risk}, we set $w_{y_1}=20$ and $w_{y_0}$ , encouraging higher sensitivity to malignant cases and aligning the optimisation with clinical settings.

\section{Results}

We evaluate Baseline, GMM Filter, Co-teaching, DivideMix and UNICON on binary DermaMNIST and Pathmnist, on 0\%, 20\% and 40\% noise rates. We assess the impact of label noise on clinical risk and the effectiveness of SOTA methods and Cost-Sensitive Loss integration. 

\subsection{Impact of Label Noise on Clinical Risk}

\begin{figure}[t]
    \centering
    \includegraphics[width=0.95\linewidth]{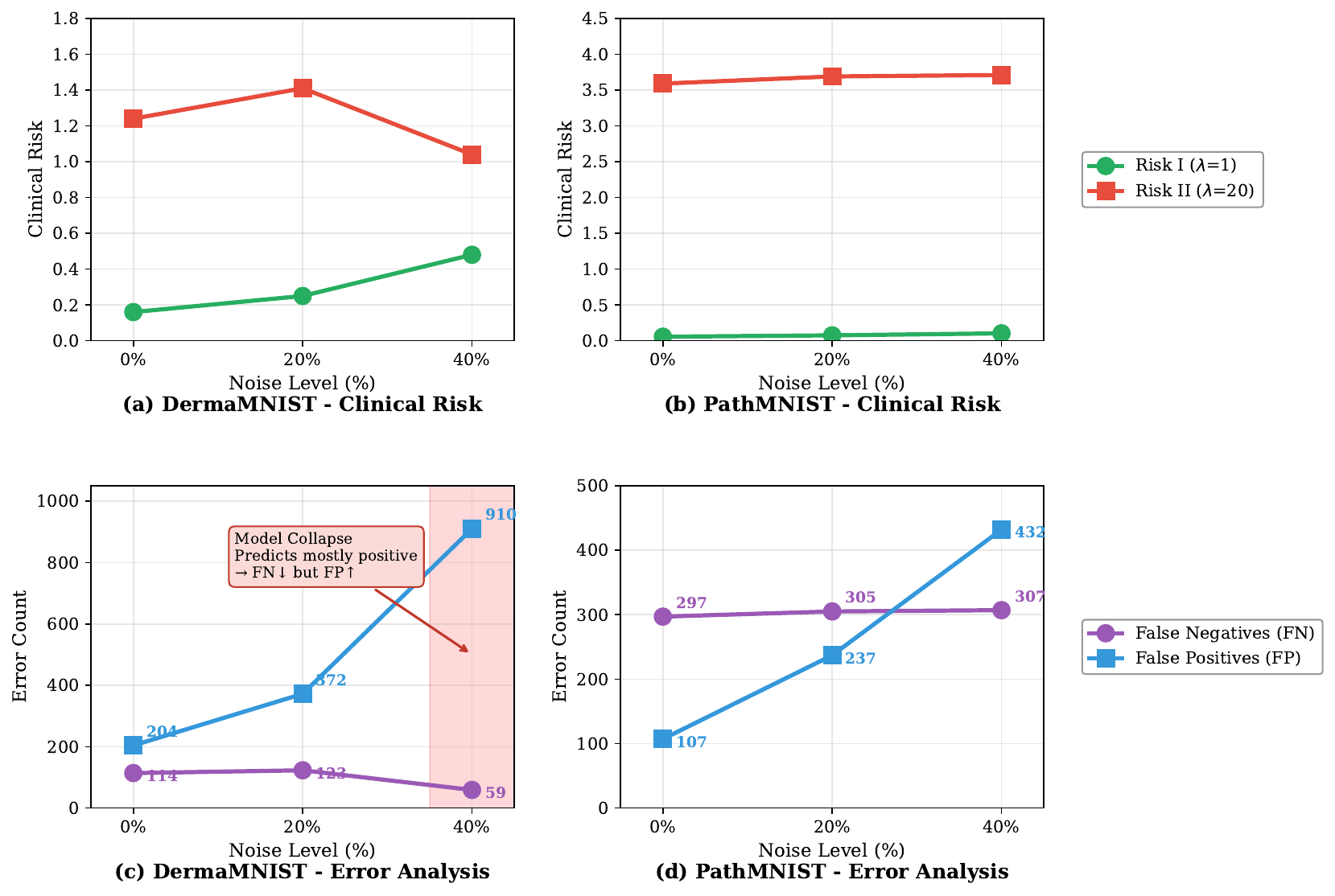}
    \caption{Impact of label noise on the Baseline model. Panels (a) and (b) show the Risk I and Risk II values across 0\%, 20\% and 40\% noise levels for DermaMNIST and PathMNIST, respectively. Panels (c) the relation of false negatives and false positives across the same settings. }
    \label{fig:noise_impact}
\end{figure}

Figure~\ref{fig:noise_impact} illustrates how label noise affects clinical risk for the Baseline model. On DermaMNIST,  Figure~\ref{fig:noise_impact}(a) and (c), we observe that Risk~I increases as the label noise get higher, but for Risk~II it decreases at 40\% noise. Error analysis reveals that the model undergoes a prediction collapse, predominantly predicting positive cases. This dramatically reduces false negatives (FN drops from 114 to 59) while false positives increase sharply (FP rises from 204 to 910). Although this artificially lowers Risk II (which heavily penalizes FN), it severely compromises diagnostic utility.

On PathMNIST, Figure~\ref{fig:noise_impact}~(b) and (d), both FN and FP increase gradually with noise, and Risk II rises monotonically, exhibiting the expected degradation pattern. The different behaviour across datasets can be attributed to PathMNIST's substantially larger training set (89,996 vs 7,007 samples), which provides greater resilience against noise-induced collapse.


\subsection{Comparison of Noise-Robust Methods}

Table~\ref{tab:results} presents comprehensive results for all methods across both datasets and noise levels. The comparison with the CS loss is represented by the `+CS` complement.

\begin{table*}[htbp]
\centering
\caption{Comparison of Baseline, GMM Filter, Co-teaching, Dividemix and UNICON across DermaMNIST and PathMNIST datasets on 0\%, 20\% and 40\% noise levels. Best values per column highlighted in gray.}
\label{tab:results}
\resizebox{\textwidth}{!}{%
\begin{tabular}{lcccccccccccccccc}
\toprule
& \multicolumn{5}{c}{\textbf{0\% Noise}} & \multicolumn{5}{c}{\textbf{20\% Noise}} & \multicolumn{5}{c}{\textbf{40\% Noise}} \\
\cmidrule(lr){2-6} \cmidrule(lr){7-11} \cmidrule(lr){12-16}
\textbf{Method} & Sens. & Spec. & BAC & AUC & F1 & Sens. & Spec. & BAC & AUC & F1 & Sens. & Spec. & BAC & AUC & F1 \\
\midrule
\multicolumn{16}{l}{\textit{DermaMNIST}} \\
\midrule
Baseline         & 0.709 & \cellcolor{gray!25}\textbf{0.874} & 0.791 & \cellcolor{gray!25}\textbf{0.897} & \cellcolor{gray!25}\textbf{0.636} & 0.686 & 0.769 & 0.728 & 0.798 & 0.521 & 0.849 & 0.436 & 0.643 & 0.688 & 0.407 \\
Baseline+CS      & 0.852 & 0.758 & 0.805 & 0.890 & 0.598 & 0.599 & 0.712 & 0.656 & 0.724 & 0.431 & 0.640 & 0.469 & 0.554 & 0.586 & 0.335 \\
\cdashline{1-16}
GMM Filter       & 0.758 & 0.818 & 0.788 & 0.889 & 0.604 & 0.758 & 0.711 & 0.734 & 0.809 & 0.514 & 0.760 & 0.591 & 0.676 & 0.717 & 0.442 \\
GMM Filter+CS    & 0.811 & 0.795 & 0.803 & 0.890 & 0.612 & 0.755 & 0.604 & 0.679 & 0.743 & 0.446 & 0.730 & 0.359 & 0.544 & 0.583 & 0.334 \\
\cdashline{1-16}
Co-teaching      & 0.755 & 0.846 & 0.800 & 0.891 & 0.632 & 0.617 & \cellcolor{gray!25}\textbf{0.857} & 0.737 & \cellcolor{gray!25}\textbf{0.843} & \cellcolor{gray!25}\textbf{0.560} & 0.508 & \cellcolor{gray!25}\textbf{0.793} & 0.650 & \cellcolor{gray!25}\textbf{0.734} & 0.430 \\
Co-teaching+CS   & 0.791 & 0.821 & \cellcolor{gray!25}\textbf{0.806} & 0.896 & 0.626 & 0.783 & 0.737 & \cellcolor{gray!25}\textbf{0.760} & 0.825 & 0.547 & 0.934 & 0.464 & 0.699 & 0.710 & 0.451 \\
\cdashline{1-16}
DivideMix        & 0.735 & 0.826 & 0.780 & 0.882 & 0.599 & 0.704 & 0.743 & 0.724 & 0.807 & 0.510 & 0.980 & 0.471 & 0.725 & 0.690 & 0.471 \\
DivideMix+CS     & \cellcolor{gray!25}\textbf{0.977} & 0.518 & 0.748 & 0.880 & 0.494 & 0.722 & 0.594 & 0.658 & 0.677 & 0.426 & 0.663 & 0.594 & 0.629 & 0.657 & 0.398 \\
\cdashline{1-16}
UNICON           & 0.735 & 0.826 & 0.780 & 0.882 & 0.599 & 0.704 & 0.743 & 0.724 & 0.807 & 0.510 & 0.980 & 0.485 & \cellcolor{gray!25}\textbf{0.732} & 0.701 & \cellcolor{gray!25}\textbf{0.478} \\
UNICON+CS        & \cellcolor{gray!25}\textbf{0.977} & 0.518 & 0.748 & 0.880 & 0.494 & \cellcolor{gray!25}\textbf{0.977} & 0.411 & 0.694 & 0.687 & 0.444 & \cellcolor{gray!25}\textbf{0.987} & 0.474 & 0.731 & 0.706 & 0.476 \\
\midrule
\multicolumn{16}{l}{\textit{PathMNIST}} \\
\midrule
Baseline         & 0.820 & 0.981 & 0.901 & 0.961 & 0.870 & 0.816 & \cellcolor{gray!25}\textbf{0.957} & 0.886 & 0.898 & 0.833 & 0.814 & 0.922 & 0.868 & 0.922 & 0.785 \\
Baseline+CS      & 0.814 & 0.980 & 0.897 & 0.962 & 0.650 & 0.758 & 0.707 & 0.732 & 0.822 & 0.554 & 0.855 & 0.224 & 0.540 & 0.649 & 0.385 \\
\cdashline{1-16}
GMM Filter       & 0.823 & 0.979 & 0.901 & 0.963 & 0.870 & 0.850 & 0.922 & 0.886 & 0.954 & 0.806 & 0.776 & 0.959 & 0.868 & 0.946 & 0.812 \\
GMM Filter+CS    & 0.832 & 0.971 & 0.902 & 0.961 & 0.863 & 0.845 & 0.892 & 0.868 & 0.943 & 0.766 & 0.855 & 0.765 & 0.810 & 0.869 & 0.648 \\
\cdashline{1-16}
Co-teaching      & 0.843 & \cellcolor{gray!25}\textbf{0.984} & \cellcolor{gray!25}\textbf{0.913} & \cellcolor{gray!25}\textbf{0.973} & \cellcolor{gray!25}\textbf{0.888} & 0.857 & 0.954 & \cellcolor{gray!25}\textbf{0.906} & \cellcolor{gray!25}\textbf{0.971} & \cellcolor{gray!25}\textbf{0.853} & 0.774 & \cellcolor{gray!25}\textbf{0.966} & \cellcolor{gray!25}\textbf{0.870} & \cellcolor{gray!25}\textbf{0.963} & \cellcolor{gray!25}\textbf{0.820} \\
Co-teaching+CS   & 0.835 & 0.971 & 0.903 & 0.961 & 0.864 & 0.865 & 0.933 & 0.899 & 0.960 & 0.828 & \cellcolor{gray!25}\textbf{0.968} & 0.192 & 0.580 & 0.759 & 0.415 \\
\cdashline{1-16}
DivideMix        & 0.829 & 0.961 & 0.895 & 0.969 & 0.463 & 0.833 & 0.933 & 0.883 & 0.949 & 0.810 & 0.728 & 0.957 & 0.842 & 0.923 & 0.777 \\
DivideMix+CS     & \cellcolor{gray!25}\textbf{0.887} & 0.905 & 0.896 & 0.952 & 0.805 & 0.628 & 0.826 & 0.727 & 0.797 & 0.569 & 0.715 & 0.799 & 0.757 & 0.798 & 0.599 \\
\cdashline{1-16}
UNICON           & 0.824 & 0.972 & 0.898 & 0.969 & 0.859 & 0.732 & 0.952 & 0.842 & 0.942 & 0.774 & 0.728 & 0.957 & 0.842 & 0.923 & 0.777 \\
UNICON+CS        & \cellcolor{gray!25}\textbf{0.887} & 0.905 & 0.896 & 0.952 & 0.805 & \cellcolor{gray!25}\textbf{0.950} & 0.617 & 0.757 & 0.791 & 0.805 & 0.904 & 0.626 & 0.757 & 0.812 & 0.599 \\
\bottomrule
\end{tabular}%
}
\end{table*}


As noise increases, methods exhibit divergent behaviors. On DermaMNIST, Co-teaching maintains high specificity but suffers in sensitivity (0.508 at 40\% noise), while DivideMix and UNICON exhibit the opposite pattern with very high sensitivity ($>$0.97) but reduced specificity, suggesting prediction collapse. On PathMNIST, Co-teaching maintains the most balanced performance across all noise levels (BAC = 0.870 at 40\%), while other methods show more modest degradation.

The integration of CS Loss amplifies existing method tendencies. UNICON+CS achieves very high sensitivity ($>$0.97) consistently but at the cost of reduced specificity. Co-teaching+CS achieves the highest BAC on DermaMNIST at 20\% noise (0.760), suggesting a more balanced sensitivity-specificity trade-off. On PathMNIST at 40\% noise, CS Loss causes instability in Co-teaching+CS (BAC drops to 0.580 due to extreme sensitivity of 0.968 with specificity of only 0.192).



\begin{figure}[t]
    \centering
    \includegraphics[width=0.95\linewidth]{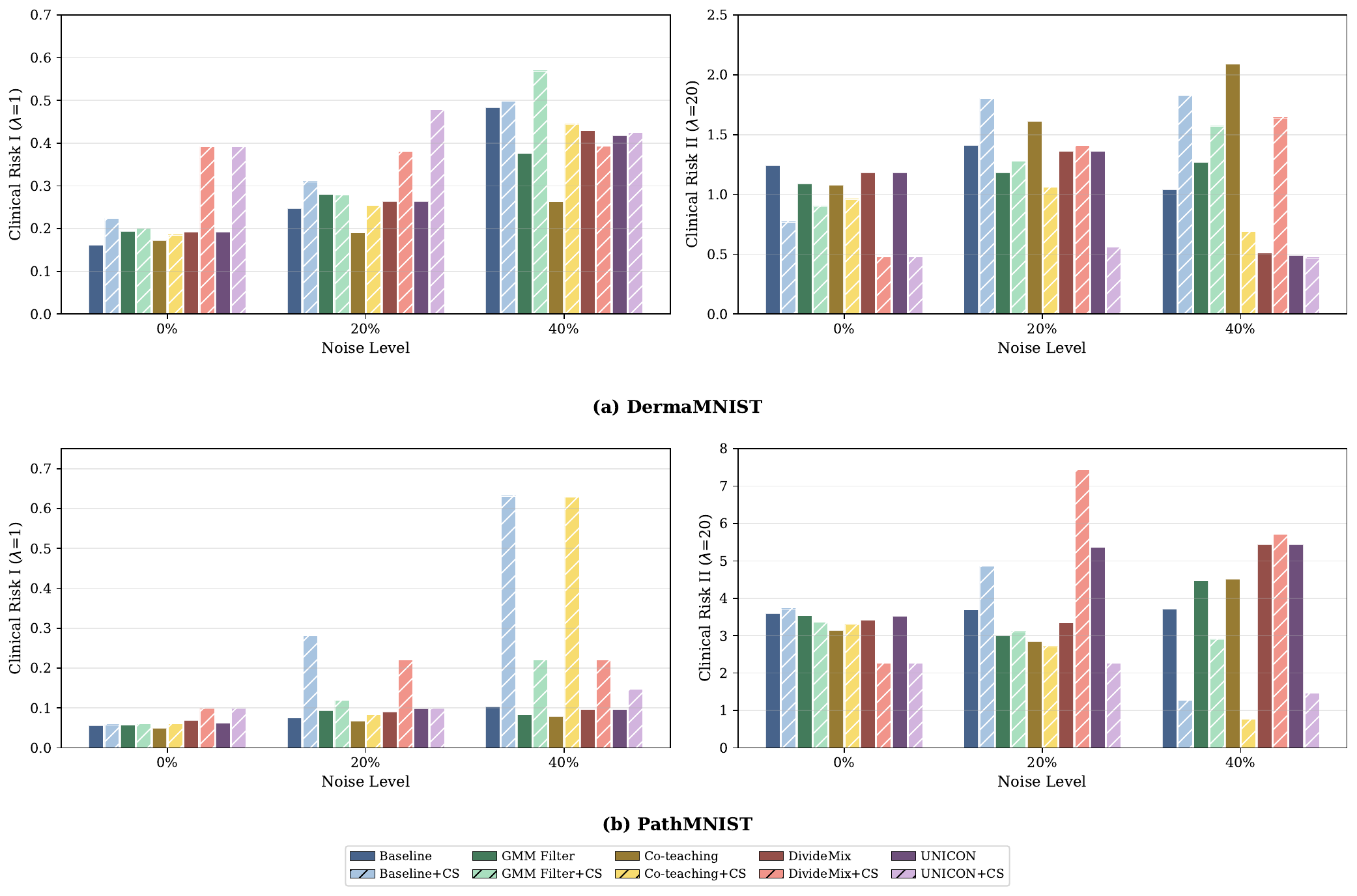}
    \caption{Clinical Risk II comparison across methods, datasets, and noise levels. Lower values indicate safer models. UNICON+CS achieves the lowest risk in most conditions.}
    \label{fig:risk_comparison}
\end{figure}

Figure~\ref{fig:risk_comparison} show Risk I and  Risk II across all noise rates, for DermaMNIST and PathMNIST. Results show that UNICON+CS consistently achieves the lowest Risk II, while for Risk I, Co-teaching achieve the lowest values. 

\begin{figure}[t]
    \centering
    \includegraphics[width=0.95\linewidth]{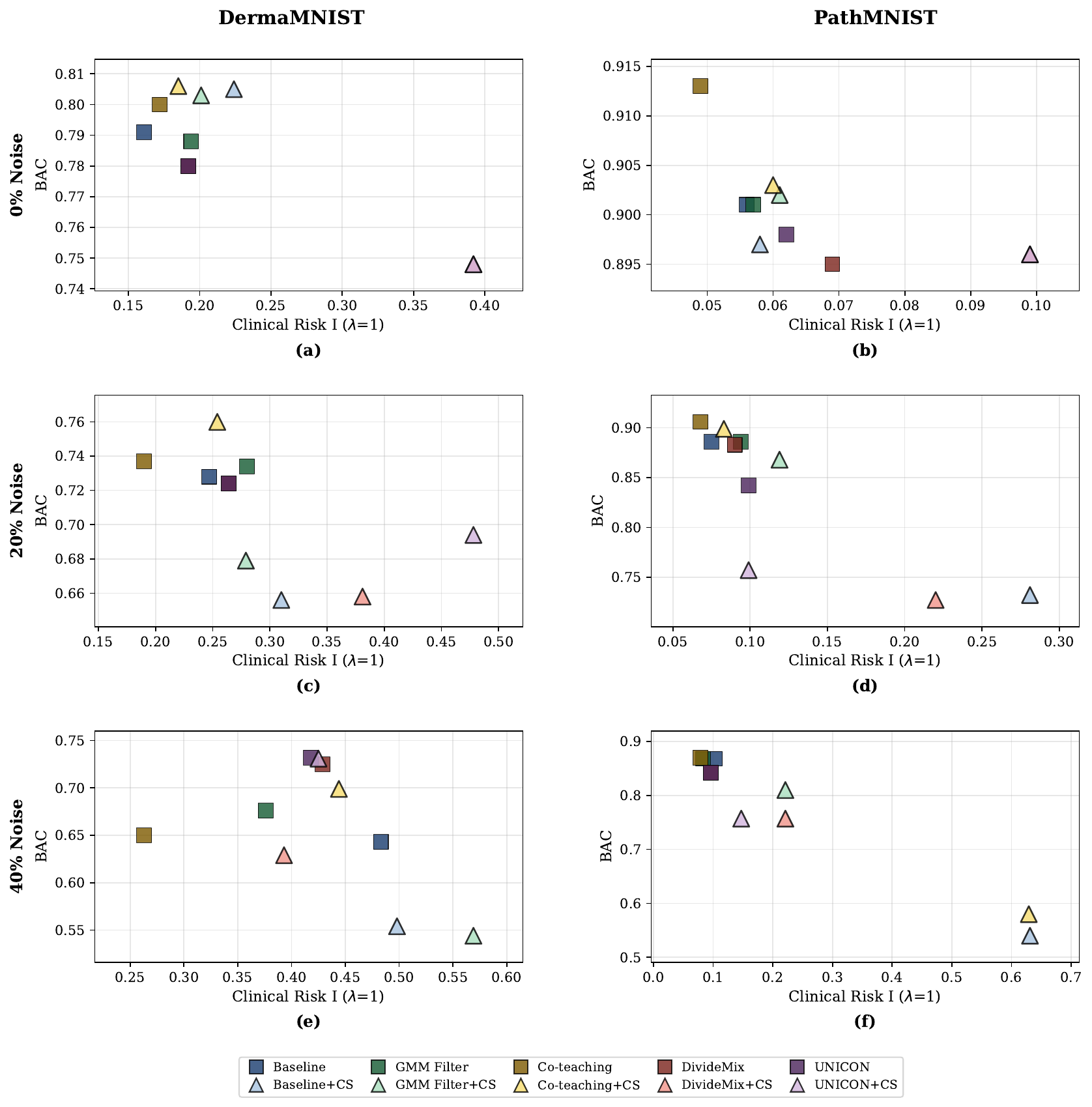}
    \caption{Trade-off between Balanced Accuracy and Clinical Risk I. The ideal operating region is the top-left corner (high BAC, low risk). }
    \label{fig:tradeoffI}
\end{figure}

\begin{figure}[t]
    \centering
    \includegraphics[width=0.95\linewidth]{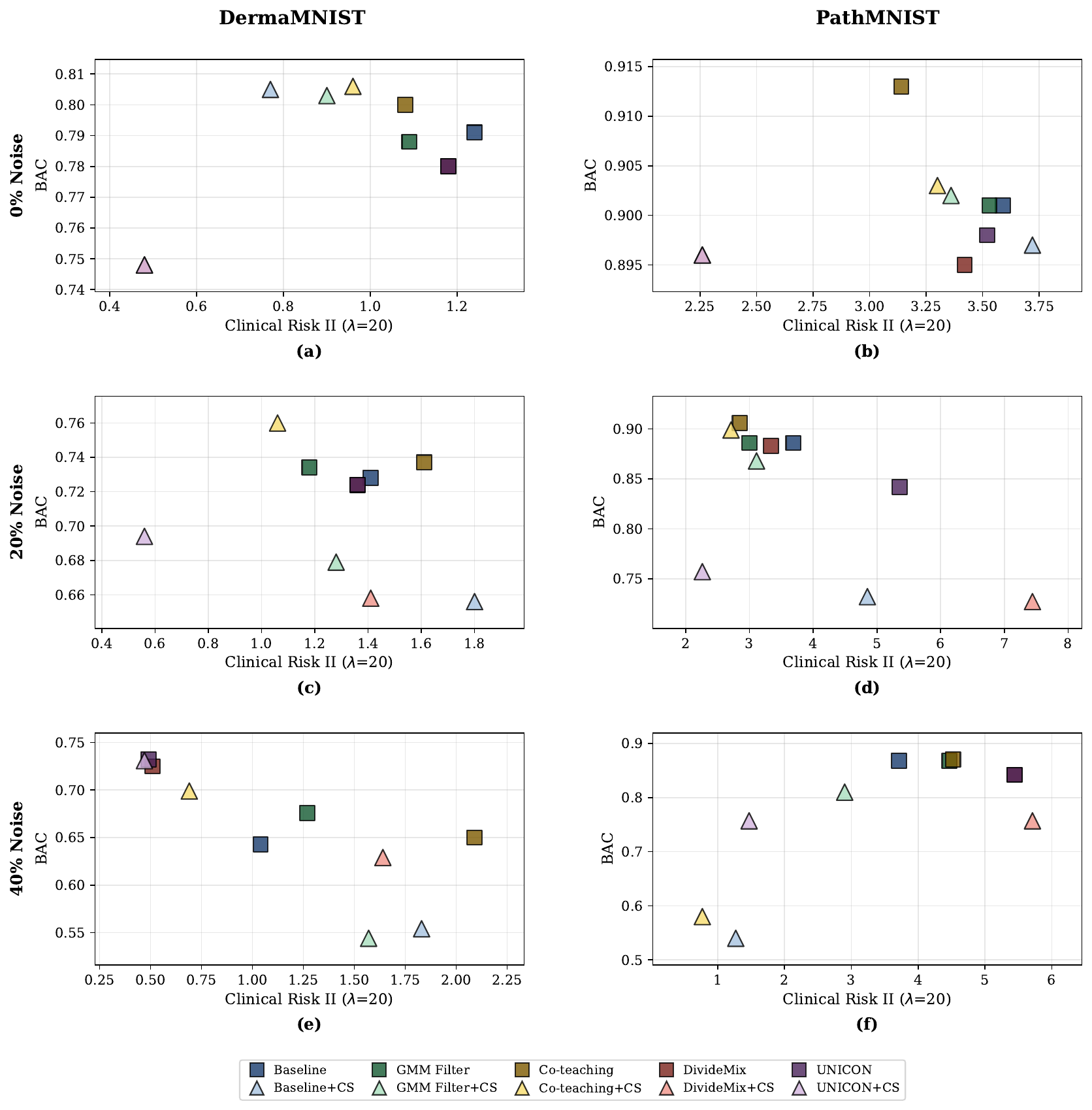}
    \caption{Trade-off between Balanced Accuracy and Clinical Risk II. The ideal operating region is the top-left corner (high BAC, low risk). }
    \label{fig:tradeoffII}
\end{figure}

Figure~\ref{fig:tradeoffI} and Figure~\ref{fig:tradeoffII}  illustrates the BAC vs. Risk I and BAC vs. Risk II trade-offs, respectivelly. From the results, we notice that Co-teaching present the one of the best BAC vs Risk II, on different noise level rates, showing a high value of BAC and low Risk I. For Risk II, showed in Figure~\ref{fig:tradeoffII}, UNICON+CS has the lower Risk II values, but mehtod with highest values of BAC deppends on the noise rate. 

We observe that the effectiveness of CS Loss is strongly method-dependent. UNICON+CS achieves consistent improvement across all conditions, with average Risk II of 0.50 on DermaMNIST and 2.00 on PathMNIST. UNICON's combination of uniform sample selection and contrastive learning appears particularly compatible with cost-sensitive optimization, as uniform selection prevents the systematic under-representation of minority class samples that can occur with other filtering strategies. Co-teaching+CS shows strong improvement, particularly at high noise (40\% on DermaMNIST), where average Risk II reduces from 1.59 to 0.90. However, it exhibits instability on PathMNIST at 40\% noise, where the model collapses toward extreme sensitivity. GMM Filter+CS yields only marginal improvement. The filtering mechanism, which uses a class-agnostic threshold of 0.5, may conflict with cost-sensitive reweighting by disproportionately filtering positive-class samples. DivideMix+CS exhibits unstable behavior, low values at 0\% noise but significantly degraded at higher noise levels, particularly on PathMNIST.
Baseline+CS shows mixed results, improving at 0\% noise but becoming unstable under higher noise conditions, confirming that cost-sensitive optimization alone is insufficient without noise-robust mechanisms.

\section{Discussion}
\label{sec:discussion}

Our results reveal that evaluating noise-robust learning solely through accuracy-oriented metrics can be misleading in medical settings with asymmetric diagnostic costs. Three key findings deserve further discussion.

First, the collapse-like behavior observed on DermaMNIST at 40\% noise illustrates a fundamental limitation of standard evaluation: Risk II can decrease even as diagnostic utility degrades, because a model that predicts almost all cases as positive trivially minimizes false negatives. This phenomenon highlights that clinical risk metrics, when used in isolation, can mask degenerate model behavior. Practitioners must therefore examine both aggregate risk values and their underlying FN/FP decomposition to ensure that low risk reflects genuine diagnostic quality rather than trivial prediction bias.

Second, the method-dependent effectiveness of CS Loss suggests that the interaction between noise robustness and cost sensitivity is non-trivial. UNICON's strong compatibility with CS Loss likely stems from its uniform sample selection mechanism, which ensures that minority class samples (malignant cases) are not disproportionately filtered during training. In contrast, methods like GMM Filter that use a single class-agnostic threshold may systematically under-select positive samples under noise, counteracting the effect of cost-sensitive reweighting. This observation motivates the development of class-aware filtering strategies that account for the asymmetric clinical costs when deciding which samples to retain.

Third, our results demonstrate that clinical risk should be treated as a first-class evaluation criterion---not merely as a post-hoc analysis---when selecting noise-robust methods for medical deployment. The substantial gap between methods that optimize BAC and those that minimize Risk II (e.g., Co-teaching achieves the highest BAC on PathMNIST at 40\% noise but has one of the highest Risk II values) underscores that practitioners need explicit risk-aware evaluation protocols.

\section{Conclusion}
\label{sec:conclusion}

In this work, we provided a clinical risk analysis of medical image classifiers trained under label noise. Beyond accuracy-oriented metrics, we evaluated noise-robust learning methods using cost-sensitive risk measures that reflect the asymmetric harm of diagnostic errors, following the Global Risk formulation.

Across DermaMNIST and PathMNIST under symmetric noise rates of 0\%, 20\%, and 40\%, we showed that: (1)~high noise can induce collapse-like behaviors that distort the FN/FP balance, producing misleadingly favorable risk scores despite degraded diagnostic utility; (2)~cost-sensitive loss integration is strongly method-dependent, with UNICON+CS achieving the lowest average Risk II (1.25), a 57\% reduction over the baseline; and (3)~methods optimizing for balanced accuracy do not necessarily minimize clinical risk, motivating risk-aware evaluation for medical AI.

Future work should investigate  adaptive cost ratios that adjust $\lambda$ based on dataset characteristics, extending this analysis to multi-class settings, and developing theoretically grounded class-aware filtering thresholds.

\bibliographystyle{sbc}
\bibliography{sbc-template}

\end{document}